\documentclass[12pt, reqno]{amsart}
\usepackage{multirow}
\usepackage{tabularx}
\usepackage{xcolor}
\usepackage{graphicx}
\usepackage[utf8]{inputenc}
\usepackage{listings}
\usepackage{tcolorbox}
\tcbuselibrary{listings}

\definecolor{codegreen}{rgb}{0,0.6,0}
\definecolor{codegray}{rgb}{0.5,0.5,0.5}
\definecolor{codepurple}{rgb}{0.58,0,0.82}
\definecolor{backcolor}{rgb}{0.95,0.95,0.92}

\lstdefinestyle{mystyle}{
  backgroundcolor=\color{backcolor},
  commentstyle=\color{codegreen},
  keywordstyle=\color{magenta},
  numberstyle=\tiny\color{codegray},
  stringstyle=\color{codepurple},
  basicstyle=\ttfamily\footnotesize,
  breakatwhitespace=false,
  breaklines=true,
  captionpos=b,
  keepspaces=true,
  numbers=none,
  showspaces=false,
  showstringspaces=false,
  showtabs=false,
  tabsize=2
}

\lstdefinelanguage{yaml}{
  basicstyle=\footnotesize\ttfamily,
  breakatwhitespace=false,
  breaklines=true,
  commentstyle=\color{red},
  identifierstyle=\color{blue},
  keywords={true,false,null,y,n},
  morecomment=[l]{\#},
  morestring=[b]',
  morestring=[b]",
  sensitive=false,
  stringstyle=\color{blue},
}

\lstdefinelanguage{csharp}{
  morekeywords={public, private, void, static, class, using, System, string, if, else, int, for, var, while, Console, WriteLine, ReadLine, double, return, new},
  morendkeywords={bool, true, false},
  sensitive=false,
  morecomment=[l]{//},
  morecomment=[s]{/*}{*/},
  morestring=[b]",
}

\newtcblisting{listingsbox}[1][]{
  listing only,
  listing options={style=mystyle},
  #1
}

\usepackage[bookmarksnumbered, colorlinks, plainpages=false]{hyperref}
\usepackage[style=numeric]{biblatex}
\begin{document}
\setcounter{page}{1}

\noindent {\small Safurai Team}\hfill     {\small Berlin, Germany}\\
{\small 6th November 2023}\hfill  {\small https://www.safurai.com}

\centerline{}

\centerline{}

\title[Safurai-Csharp]{Safurai-Csharp: Harnessing Synthetic Data to improve language-specific Code LLM}

\author[Safurai Team]{Cifarelli D., Boiardi L., Puppo A., Jovanovic L.}

\begin{abstract}
This paper introduces Safurai-Csharp, an open-source model designed to specialize in the generation, completion, and debugging of C\# code. Safurai-Csharp is built upon the novel CodeLlama 34B model and leverages the EvolInstruct technique, creating a refined and expanded dataset for its fine-tuning process. The results of its performance, a notable score of 56.33\% on the Manual MultiPL-E benchmark (Zero-Shot, Pass@1), signal its high capacity to streamline developers' workflows and aid code learning. It shows promise in setting new stakes in the landscape of open-source C\# LLMs and hopes to inspire more inclusive and wide-ranging development in the field of language-specific LLMs. 
\end{abstract} \maketitle

\section{Introduction}
Artificial intelligence's influence on software development and coding is immense and rapidly evolving. OpenAI models have led the closed-source models landscape; the open-source community has risen to the challenge primarily with models like Starcoder\cite{starcoder}, WizardCoder\cite{wizardcoder} and PanguCoder\cite{pangucoder}. While these models have significantly closed the performance gap between open and closed-source models in general, the need for language-specific large language models remains apparent. Safurai-Csharp, introduced to fill this gap, excels in programming for the C\# language, offering a significant milestone in AI-driven coding tools.  

The recent introduction of CodeLlama\footnote{https://ai.meta.com/research/publications/code-llama-open-foundation-models-for-code} by Meta also imposes new potential foundational models for Coding LLMs, marking another significant stride in this burgeoning area. Safurai-Csharp aims to capitalize on these cutting-edge foundational models and novel techniques to offer competitive solutions in the realm of code generation specifically for C\#. 

This paper presents an open-source model – Safurai-Csharp, which specializes in facilitating code completion, debugging, and generation for the C\# programming language. With a remarkable 56.33\% score in the Manual MultiPL-E benchmark (Zero-Shot, Pass@1), Safurai-Csharp exhibits a strong capability to generate code and natural language about code, streamlining developers' workflows and assisting in code learning. 

The model achieves this feat by leveraging a rich and efficiently filtered dataset of 13 billion tokens specific to the C\# language. It incorporates the novel Evol Instruct method, introduced by Xu et al. in 2023 with the WizardCoder LLM\cite{wizardcoder}, which enriches and diversifies the initial instruction dataset through iterative evolution and upgrading processes. 

In summary, our contributions in this paper include:

\begin{itemize}
    \item Introducing Safurai-Csharp, a state-of-the-art model specialized for C\# language operations, including code completion, debugging, and generation.
    \item Showcasing remarkable performance with a 56.33\% score in the Manual MultiPL-E benchmark (Zero-Shot, Pass@1), setting a new standard for open-source C\# language model.
    \item Implementing the creation of the synthetic dataset through the data distilliation of CodeLlama-34b combined with the Evol Instruct technique, to iteratively enrich and diversify the initial C\# instruction dataset. 
\end{itemize}

We hope that Safurai-Csharp provides a robust opportunity for various sectors and adds a new valuable model to the Open-Source variety of Coding LLMs.

\section{Related Work}

\subsection{Coding Large Language Models}
The Codex, a 12 billion parameters model, signifies a remarkable breakthrough in the realm of Python coding, successfully solving roughly 72\% of Python-based tasks. This groundbreaking achievement has propelled the advancement of other evolved code producing models such as AlphaCode \cite{alphacode}, PaLM Coder \cite{palm}, and PanGu-Coder \cite{pangucoder}. Nevertheless, these cutting-edge models' lack of open-source availability has been a significant limitation. This gap, however, has been addressed with the introduction of numerous open-source alternatives such as CodeParrot, PolyCoder, PyCodeGPT, SantaCoder \cite{santacoder}, and StarCoder \cite{starcoder}, rejuvenating the code generation discipline. Additionally, the continual broadening of the application arenas for code generation attests to the field's escalating practical relevance. For instance, CodeGeeX \cite{codegeex}, BLOOM \cite{bloom} and ERNIE-Code \cite{erniecode} were specifically developed to facilitate multilingual modeling. Moreover, JuPyT5 \cite{jupyt5} was trained on a vast corpus of Jupyter notebooks, targeting the enhancement of interactive programming. In an effort to infuse language models with API functionality, models such as DocCoder and APICoder \cite{apicoder} have also been created. Furthermore, models like InCoder \cite{incoder}, SantaCoder, and StarCoder provide support for code generation in diverse locations. In recent times, some groups have exploited instructional tuning techniques to access the enormous potential expertise encapsulated within large language models. This procedure involves diligent refinement of these models using high-quality datasets. In the sphere of code generation, models such as WizardCoder (15B), PanguCoder and phi-1 (1.3B) distinguished themselves through exceptional performance, achieved through meticulous fine-tuning using data produced by OpenAI’s GPT-3.5 and GPT-4.

In this context, it is pertinent to mention the CodeLlama \footnote{https://ai.meta.com/research/publications/code-llama-open-foundation-models-for-code} models developed by Meta AI. Code Llama is a subsequent training iteration of Llama 2, specifically designed for code generation and completion. It also offers code debugging and supports a plethora of programming languages including Python, C++, Java, PHP, Typescript (Javascript), C\#, Bash, and more. Code Llama variants are available in three sizes, i.e., 7B, 13B and 34B parameters, each trained with 500B tokens of code and code-related data. These models cater to diverse requirements, with the 7B model being deployable on a single GPU, while the 34B model, although requiring more resources, provides superior results. Furthermore, Code Llama has two specialized versions: Code Llama – Python, which is specifically fine-tuned on Python code, and Code Llama – Instruct, which focuses on user instruction comprehension for code generation.

From Microsoft Research, Phi1.5 \cite{phi-1.5} is a notable development, as an evolutionary step from Phi1 \cite{phi-1}. The Phi-1.5 model leverages Large Language Models (LLMs) to generate top-quality, textbook-like datasets. This model retains the architectural design of its predecessor, Phi-1, but improves upon it by training on a dataset consisting of 30 billion tokens, of which approximately 20 billion tokens were artificially synthesised from GPT-3.5. The outcomes of Phi-1.5 are comparable with models five times its size on typical language tasks, and it surpasses most LLMs in more complex reasoning assignments. Additional versions of Phi-1.5 were developed using traditional filtered web data along with synthetically created NLP data, named phi-1.5-web-only and phi-1.5-web, to further leverage the potential of this model.

Also the Phind\footnote{https://www.phind.com/blog/code-llama-beats-gpt4} Models reflect an unprecedented evolution in the field of Coding LLMs, demonstrating remarkable capabilities to surpass the proficiency exhibited by GPT-4 on HumanEval. Meta AI developed and fine-tuned two versions of CodeLlama-34B, i.e., CodeLlama-34B and CodeLlama-34B-Python, on an exclusive Phind dataset. Both models demonstrated exemplary performance, achieving 67.6\% and 69.5\% pass@1 on HumanEval, respectively. It is noteworthy that these figures are superior to GPT-4's performance, which reached 67\% as per the official technical report released in March. These refined models were tested using OpenAI's decontamination methodology to confirm the validity of the results. Recently, the training of a new model named Phind-CodeLlama-34B-v2 resulted in further augmentation of pass@1 score, reaching 73.8\% on HumanEval. These achievements reaffirm the potential of continuous learning and fine-tuning in enhancing the efficacy of Code LLMs, setting new benchmarks in the field.

In the midst of this exciting ongoing development in the Coding LLMs landscape, a model that struck an effective balance between the essentials of performance and efficiency emerged. Named as Mistral 7B\cite{mistral}, this model dispels the prevalent notion that augmenting model performance mandates a proportionate increase in the model size. Not only does Mistral 7B present an efficient alternative to this scalibility issue, it demonstrates that strategic design and thoughtful training can yield high performance while ameliorating computational burdens and inference latency. Exceptionally remarkable about this model is that despite belonging to the 7 billion parameter category, it managed to outclass its larger counterparts such as the 13 billion parameter Llama 2\cite{llama2} across all tested benchmarks, and even outstripped the 34 billion parameter Llama 34B in mathematics and code generation tasks. 

\subsection{Synthetic data generation through distillation}
Synthetic data generation through distillation is a machine learning technique that seeks to create artificial datasets that mimic the underlying structures and complexities of real-world data. Essentially, what distillation does is to create a model (teacher model) that can generate data, which feels 'real' but is actually artificial. This data can then be used to train another model (student model), which ideally, will learn to generalize better from the diversified data produced by the teacher model.

Microsoft Research introduced a model known as Phi-1\cite{phi-1}, which is revolutionary in terms of using synthetic data created through distillation for model training. Phi 1 criticizes the usage of standard source of code generation such as The Stack and web-based datasets like StackOverflow and CodeContest, arguing they lack the necessary properties for efficient understanding and planning of algorithms. Phi 1 finds the widely used datasets to be fragmented, containing trivial or boilerplate code, and having an unbalanced distribution of coding concepts and skills. Consequently, Microsoft Research introduced an innovative training technique that utilizes three major datasets namely a filtered code-language dataset, a synthetic textbook dataset and a small synthetic exercises dataset.

A significant part of their training data comprises less than 1 billion tokens of synthetic data derived from GPT-3.5 generated Python textbooks. This innovative synthetic data generation technique through distillation allowed Microsoft Research to develop superior models like Phi 1 and its evolved version, Phi 1.5\cite{phi-1.5}. The Phi-1.5 model utilizes Large Language Models (LLMs) to generate top-quality, textbook-like datasets for its training. It was trained on a dataset of 30 billion tokens, where approximately 20 billion tokens were synthetically created from GPT-3.5. The Phi-1.5 model displayed impressive results, matching the efficiency of larger models five times its size on regular language tasks, and outperforming most LLMs on more complex reasoning assignments.

\subsection{Data Augmentation Techniques}
Data augmentation in the context of LLMs fine-tuning refers to various techniques used to increase the size and improve the quality of training data. This is done through making certain modifications to the existing data which could entail tasks like synonyms replacement, sentence restructuring, translation between multiple languages, adding noise to data, and others. The primary goal is to create a robust model that understands diverse patterns with increased accuracy. This process helps in minimizing over-fitting, improving the model's performance and its ability to generalize over unseen data.

An intriguing example of data augmentation techniques specifically tailored for coding tasks in LLMs can be seen in Microsoft's application of the EvolInstruct method for finetuning the WizardCoder\cite{wizardcoder} model. This method revolves around a unique three-stage pipeline that progressively evolves complex instructions, refines the respective responses, and then eliminates deficient instruction data. 

The EvolInstruct approach dives deep into upscaling the complexity of instructions through a dual strategy - 'In-depth Evolving' and 'In-breadth Evolving'. The former method enhances instruction complexity by integrating constraints, deepening contextual information, concretising abstract ideas, increasing the chain of reasoning steps, and complicating input data. Conversely, the 'In-breadth Evolving' strategy is designed to create a multitude of new instructions deriving from the existing ones, thereby fostering an expansively diverse set of commands.

Empirical evidence validates the effectiveness of the EvolInstruct method, as it was meticulously deployed for finetuning the LLaMA LLM, leading to the development of the proficient WizardLM model. 

\subsection{Fine-tuning with LoRA and QLoRA}
LoRA\cite{lora} (Low-Rank Adapters) and QLoRA (Quantized LoRA) are revolutionary finetuning techniques for Large Language Models (LLMs). These techniques serve as efficient and practical solutions for managing the large, resource-intensive nature of LLMs. With conventional LLMs, their vast storage requirements and heavy GPU utilisation limits their scalable applications. However, LoRA and QLoRA manage these limitations remarkably, enabling the fine-tuning of the model while reducing significant storage and computational costs.

LoRA offers a strategic approach by introducing the concept of low-rank approximations of the changes that are made to the weight matrices during training. Instead of updating the complete weight matrix, the LoRA approach approximates these changes and stores them separately. These changes can be applied later when needed for a specific task, in essence, making them adaptable lenses or attachments to the larger model. However, where LoRA really shines is in the immense memory savings enabled through matrix decomposition, where large matrices are decomposed into smaller matrices, reducing the overall data requirements.

QLoRA\cite{qlora}, on the other hand, takes a different yet complementary approach to the problem. It introduces the concept of quantization, where the model weights are compressed into a smaller data format, significantly reducing the memory footprint. QLoRa is essentially a combination of quantization and LoRA. The model is first quantized into a smaller data format, then LoRA training is performed, reducing further the computational load. 

In a nutshell, LoRA and QLoRA are transformative techniques that not only address the major challenges associated with managing and deploying LLMs but also bring finetuning a step closer to practicality. They ensure high performance of the models, reducing not only storage cost but also computational power requirements. This ability to reduce resources while maintaining model performance establishes LoRA and QLoRA as significant advancements in the machine learning field.

\section{Methods}

\subsection{Initial dataset}
This section is dedicated to the detailing of our computational process in creating our initial dataset which we used for the fine-tuning of the LLM.

The foundation model, the CodeLlama34B\footnote{https://huggingface.co/codellama/CodeLlama-34b-hf}, designed by Meta AI, played an instrumental role in the genesis of the dataset. CodeLlama34B, with its remarkable capacity to generate programming problems, was utilized to produce an initial batch of 100,000 examples of C\# programming tasks. The tenacity of the model's output provided us a large pool from which we could sift and gather top-tier examples that best suited our needs for optimization and refinement.

However, merely generating a large number of examples was not sufficient; a critical step that followed was the filtration of this dataset. Not all problems generated were required or ideally suited for our particular training set. For instance, we encountered issues with tasks that were too analogous, presenting no significant variance that would offer productive learning for our LLM. Similarly, certain tasks were excessively verbose, which though useful in certain contexts, did not align with the objectives outlined for Safurai-Csharp.

Thus, the steps taken to purify the dataset involved removing these undesirable examples until we were left with a collection of 4,000 problems—one that was both rich in instructional value and diversity, and manageable in terms of size.

Further enrichment of the dataset was achieved by incorporating solutions for each of these C\# problems. The same engine, CodeLlama34B, was used to generate relevant solutions for each problem. This not only provided a cohesive dataset of problem-solution pairs, but also ensured comprehension and functionality across a range of C\# tasks, thereby enhancing the learning scope of our LLM.

\subsubsection{The distillation prompt used to generate 100,000 examples from Codellama34B}
\
\begin{listingsbox}[listing options={language=Python}]
synthetic_seed_csharp = ("""
    Write 1000 programming interview questions of easy and medium complexity.
   Provide questions on a diverse range of subjects and make sure no two questions are alike.
   Be creative and think about a lot of different topics with different difficulties.
   The question needs to be specific to csharp language.
   Enumerate all the questions.
   """)
\end{listingsbox}

\subsection{Data Augmentation with Evol-Instruct}
We employed the EvolInstruct process, a technique notably applied by Microsoft in their fine-tuning of the WizardCoder model. Our dataset, initially comprising 4,000 pairs of C\# problems and solutions, underwent augmentation with the application of EvolInstruct thrice, culminating in a significantly expanded and diversified dataset of 16,000 pairs.

The application of EvolInstruct to our dataset adhered to its three-stage pipeline, evolving complex instructions initially, followed by refining associated responses, and, lastly, eliminating unsatisfactory instruction data. 

The core of this methodology rests on enhancing the complexity of instructions—the fundamental element in our problem dataset. Here, the 'In-depth Evolving' strategy was extensively used, interweaving constraints, cementing abstract concepts, deepening contextual information, iterating the reasoning steps, and complicating input data, resulting in a series of intricate and diverse programming tasks. 

The augmentation process also included the 'In-breadth Evolving' strategy, which led to proliferation of our problem set, ensuring an array of diverse instructions stem from the existing ones. The result was a template with manifold complex tasks, embodying a much broader set of commands.

This approach to data augmentation through the EvolInstruct method proved to be beneficial in optimizing our LLM. Owing to its documented success in fine-tuning the WizardCoder model, its application to our dataset was apt and resulted in a significant improvement in the quality of the training data, bolstering the performance of our Safurai-Csharp model.

\begin{figure}[h]
  \centering
  \includegraphics[width=0.8\textwidth]{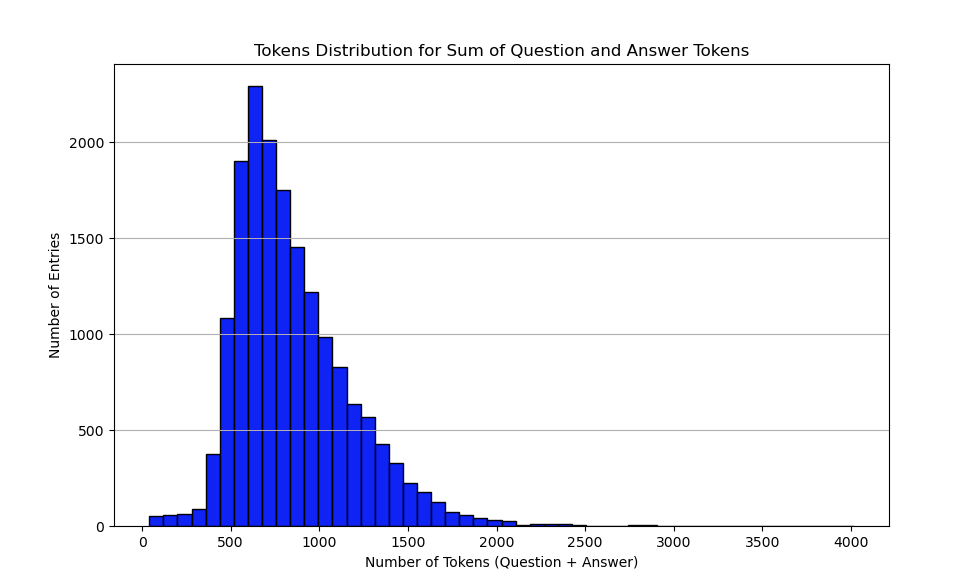}
  \caption{Token distribution in our dataset}
  \label{fig:grafico}
\end{figure}

\subsection{Fine-tuning Setup}
Here is a walkthrough of how we finetuned the Safurai-Csharp model.
To begin with, the base model and its respective configuration were implemented using CodeLlama-34b-hf. This base model was derived from the CodeLlamaForCausalLM model class and employed the CodeLlamaTokenizer type, which is specifically designed for programming languages.
The model was set to load in 4-bit mode to save memory and increase the processing speed. The "strict" mode, used to stringently match the model’s state dictionary with the loaded state dictionary, was set to 'false' to allow smooth loading even if certain keys are missing.
Safurai-Csharp was trained on the dataset we previously described. The model adopts sample packing and adjusts the input to a sequence length of 4096 to optimize GPU memory usage.

One of the distinguishing features of this finetuning process is the use of Low Rank Adapters (LoRA). By setting the 'adapter' option to ‘lora’, we introduced efficient layer-wise low-rank adapter training. We further customized the LoRA setting with a rank (lora\_r) of 32 and a scaling factor (lora\_alpha) of 16. These settings served to cap the complexity of LoRA while preserving the most significant components from the full-rank space. A dropout rate of 0.05 was set to prevent overfitting during the finetuning phase.

Training parameters were specified to ensure model efficiency. Gradient accumulation steps were set to 4, with a micro-batch size of 2 and three epochs to cover computations for a learner over the entire dataset.
To optimize the model, we went with the adamw\_bnb\_8bit optimizer. Cosine learning rate scheduling was used, and the initial learning rate was set to 0.0003. 

Gradient checkpointing was enabled for better GPU memory usage when dealing with large models and long sequences. Advanced transformer optimizations were achieved through 'flash\_attention' set as true.
The model was designed to perform regular validation checks after 40 steps, also known as warmup steps. The logging of training dynamics was set to occur after every step.

Lastly, with respect to special tokens, the model used \texttt{<s>} for beginning of stream, \texttt{</s>} for end of stream, and \texttt{<unk>} for unknown tokens. 

In summary, the Safurai-Csharp model has been created with careful selection of parameters and techniques like Low-Rank Adapters (LoRA) and Quantization, ensuring efficient finetuning of the LLM.

\begin{figure}[h]
  \centering
  \includegraphics[width=0.7\textwidth]{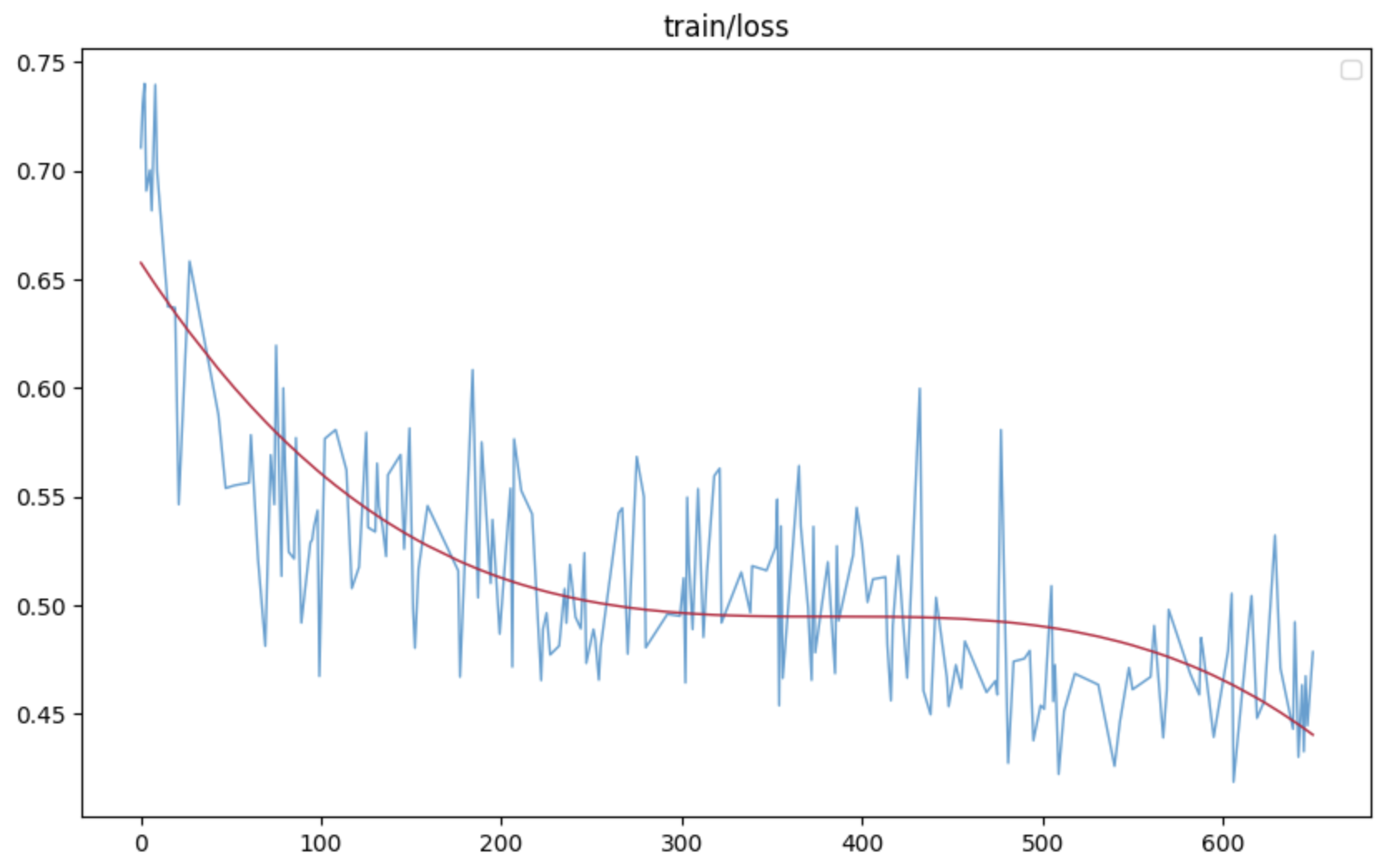}
  \caption{Train/Loss Graph}
  \label{fig:trainloss}
\end{figure}
\newpage
\subsubsection{The YAML configuration file for Fine-Tuning}
\
\begin{listingsbox}[listing options={language=yaml}]
base_model: codellama/CodeLlama-34b-hf
base_model_config: codellama/CodeLlama-34b-hf
model_type: LlamaForCausalLM
tokenizer_type: CodeLlamaTokenizer
is_llama_derived_model: true
hub_model_id: "Safurai/Evol-csharp-v1"

load_in_8bit: false
load_in_4bit: true
strict: false

datasets:
  - path: Safurai/EvolInstruct-csharp-16k-13B-Alpaca
    type: alpaca
dataset_prepared_path: last_run_prepared
val_set_size: 0.01
output_dir: ./qlora-out

sequence_len: 4096
sample_packing: true
pad_to_sequence_len: true

adapter: lora
lora_model_dir:
lora_r: 32
lora_alpha: 16
lora_dropout: 0.05
lora_target_linear: true
lora_fan_in_fan_out:

wandb_project: codellama-csharp
wandb_entity:
wandb_watch:
wandb_run_id:
wandb_log_model:

gradient_accumulation_steps: 4
micro_batch_size: 2
num_epochs: 3
optimizer: adamw_bnb_8bit
lr_scheduler: cosine
learning_rate: 0.0003
\end{listingsbox}
\
\begin{listingsbox}[listing options={language=yaml}]
train_on_inputs: false
group_by_length: false
bf16: true
fp16: false
tf32: false

gradient_checkpointing: true
early_stopping_patience:
resume_from_checkpoint:
local_rank:
logging_steps: 1
xformers_attention:
flash_attention: true

warmup_steps: 40
eval_steps: 40
save_steps:
debug:
deepspeed:
weight_decay: 0.0
fsdp:
fsdp_config:
special_tokens:
  bos_token: "<s>"
  eos_token: "</s>"
  unk_token: "<unk>"
\end{listingsbox}

\section{Experimental Results}

\subsection{Manual MultiPL-E Evaluation Results}
HumanEval\footnote{{https://github.com/openai/human-eval}}, established by OpenAI, has been the go-to benchmark in evaluating coding AI models. It focuses on the model's code-generation capacity based on precise requests and provides standard solutions and tests. MultiPL-E\footnote{https://huggingface.co/datasets/nuprl/MultiPL-E/} extends the boundaries of the current benchmarks by translating them to new languages and thereby becoming a massive multi-language benchmarking platform. It is continuously expanding in terms of number of programming languages, providing a useful comparison point for models like Safurai-Csharp.

Our model, Safurai-Csharp, achieved a pass@1 score (Zero-Shot) of 56.33\% on the Manual MultiPL-E Benchmark for C\# with n=5. Looking beyond the conventional scopes of evaluation, our team set out on the task of MultiPL-E "custom" evaluation, leaving no stone unturned to extract the most accurate measure of our model, Safurai-Csharp's performance. Basing ourselves on the MultiPL-E dataset, we conducted tests involving the 158 question-prompts, dedicating 5 unique answer to each prompt.

Traditionally, the MultiPL-E benchmark is designed to take just one function from the Safurai response to complement the function specified in the question-prompt. Safurai very often use helper functions in its responses. So, our team adopted manual evaluations (we completed the code and compiled it to have tests results) to substantiate the true test score and compare it with Codellama 34B\footnote{https://huggingface.co/codellama/CodeLlama-34b-hf}, Mistral 7B Instruct\footnote{https://huggingface.co/mistralai/Mistral-7B-Instruct-v0.1}, GPT4\cite{gpt4} and ChatGPT (we manually tested also these models in this process).

\begin{table}[h]
\centering
\caption{Results of pass@1(\%), Zero-Shot, n=5 on manual MultiPL-E C\# Benchmark}
\label{tab:performance}
\renewcommand{\arraystretch}{1.5}
\scalebox{0.8}{
\begin{tabularx}{\textwidth}{ 
  >{\hsize=1.5\hsize\centering\arraybackslash}X
  >{\hsize=.5\hsize\raggedright\arraybackslash}X 
  >{\hsize=.5\hsize\centering\arraybackslash}X}  
 \hline
  \textbf{Model} & \textbf{Size} & \textbf{C\#} \\
 \hline
 \multicolumn{3}{c}{\textbf{Closed source models}} \\
 \hline
GPT-3.5 Turbo [2023] & - & 58.22 \\
\hline
GPT-4 [2023] & - & 65.19 \\
\hline
\multicolumn{3}{c}{\textbf{Open source models}} \\ 
\hline
Mistral-7B-Instruct-v0.1 [2023] & 7B & 7.59 \\
\hline
Codellama 34B [2023] & 34B & 30.4 \\
\hline
\textbf{Safurai-Csharp [2023]} & \textbf{34B} & \textbf{56.33} \\
\hline
\end{tabularx}
}
\end{table}

\subsection{Improvements from our Foundation Model}

The evolution from our foundation model, CodeLlama, to Safurai-Csharp stands as a testament to the power of fine-tuning and optimized dataset curations. Safurai-Csharp's performance as measured by the Manual MultiPL-E Benchmark\footnote{https://huggingface.co/datasets/nuprl/MultiPL-E}, scored an impressive 56.33\%, almost double that of its predecessor, CodeLlama, which scored 30.4\%. This marked improvement is a clear reflection of the considerable strides made during the fine-tuning process.

A key contribution that led to Safurai-Csharp's impressive performance gain was the application of the EvolInstruct technique in the augmentation and crafting of the final dataset. Not only did this method help in creating a robust and diverse array of C\# problems, but it also contributed greatly to Safurai-Csharp's ability to generalize and solve unseen tasks effectively. The resulting superior performance of Safurai-Csharp illustrates the virtue of careful and meticulous dataset preparation and the value of elaborate fine-tuning processes.

\subsection{Comparison of Safurai-Csharp, Codellama 34B, ChatGPT, GPT-4 and Mistral-7B-Instruct-v0.1 on MultiPL-E C\# First Problem}

\begin{center}
    \textbf{Problem}
\end{center}

\begin{listingsbox}[listing options={language=csharp}]
    from typing import List

using System; 
using System.Numerics; 
using System.Diagnostics; 
using System.Collections.Generic; 
using System.Linq; 
using System.Text; 
using System.Security.Cryptography; 

class Problem { 
    // Return length of given string 
    // >>> StringLength(("")) 
    // (0L) 
    // >>> StringLength(("abc")) 
    // (3L) 
    public static long Strlen(string str) {
        //code here
    }
}
\end{listingsbox}

\begin{center}
    \textbf{Safurai-Csharp Output}
\end{center}
Sure! Here's a optimized solution to the `StringLength` function in C\#:

\begin{listingsbox}[listing options={language=csharp}]
from typing import List
using System;

class Program {
    // Return length of given string
    public static long StringLength(string str) {
        // Use the built-in Length property to get
         the length of the string
        return str.Length;
    }
\end{listingsbox}

\begin{listingsbox}[listing options={language=csharp}]
    static void Main(string[] args) {
        // Test the StringLength function
        Console.WriteLine(StringLength(""""));  
// Output: 0
        Console.WriteLine(StringLength(""abc""));  
// Output: 3
    }
}
\end{listingsbox}
Explanation:

1. The `StringLength` function takes a string `str` as input and returns its length as a `long` value.
2. To get the length of a string, we can use the `Length` property of the `string` class, which provides the number of characters in the string.
3. In the `Main` method, we test the `StringLength` function by passing two sample inputs and printing their lengths.

This solution is optimized as it uses the built-in functionality of the `string` class, which provides an efficient way to get the length of a string. The code is easy to read and understand due to its simplicity and clear comments."
\begin{center}
    \textbf{ChatGPT Output}
\end{center}
It looks like you've started defining a C\# class called Problem with a Strlen method that is supposed to return the length of a given string. However, the code snippet you provided is incomplete. I'll help you complete the Strlen method and add a sample Main method to demonstrate how to use it:

\begin{listingsbox}[listing options={language=csharp}]
from typing import List
using System;

class Problem
{
    // Return length of given string
    public static long Strlen(string str)
    {
        return str.Length;
    }
\end{listingsbox}
\begin{listingsbox}[listing options={language=csharp}]
    public static void Main(string[] args)
    {
        string input1 = ""; // Empty string
        string input2 = "abc";

        long length1 = Strlen(input1);
        long length2 = Strlen(input2);

        Console.WriteLine("Length of input1: " + length1); 
// Should print 0
        Console.WriteLine("Length of input2: " + length2); 
// Should print 3
    }
}
\end{listingsbox}
In this code, I've completed the Strlen method by using the Length property of the string to return its length. I've also added a Main method where you can see how to use the Strlen method to calculate the lengths of different strings.

\begin{center}
    \textbf{GPT-4 Output}
\end{center}
Here is the C\# implementation for Strlen function which is used to return the length of the given string.

\begin{listingsbox}[listing options={language=csharp}]
public static long Strlen(string str) {
    return str.Length;
}
\end{listingsbox}

In the function above, the Length attribute of "string" type is used. This attribute returns the number of characters in the current string.

\begin{center}
    \textbf{Mistral 7B Instruct Output}
\end{center}
Here's a possible implementation for the StringLength method in C\#:

\begin{listingsbox}[listing options={language=csharp}]
public static long Strlen(string str) {
    return str.Length;
}    
\end{listingsbox}

This method simply returns the length of the input string as an integer value, which is then cast to a long. The str.Length property provides access to the number of characters in the string.

\section{Conclusion}

The rapid expansion of machine learning and AI in the field of programming has necessitated the urgent need for language-specific LLMs. Our contribution through Safurai-Csharp is an experimental testament to the capabilities of such models in better understanding and generating code in a nuanced programming language like C\#. Boasting a remarkable score in the Manual MultiPL-E benchmark, Safurai-Csharp underscores the impressive advances that can be achieved with dedicated tools and strategies like the Evol Instruct technique for dataset enhancement and diversification.

Furthermore, through the fine-tuning of the advanced foundational model, CodeLlama, Safurai-Csharp has exemplified how performance strides can be made in large language models, potentially serving as a blueprint for future endeavors focusing on other languages. The open-source nature of Safurai-Csharp holds promises for broader inclusivity, accessibility, and further contributions from the global community, aligning well with the collaborative spirit of AI research and development.

The richness and versatility of Safurai-Csharp's capabilities, demonstrated in code completion, debugging, and generation for the C\# programming language, makes it a compelling tool that could revolutionize developers' workflows and the journey of code learners. We strongly believe that this open-source initiative is a monumental stride in language-specific LLMs, bringing us closer to real-world applicability and integration of AI in coding.

We envisage the continued evolution and refinement of such specialized coding LLMs and eagerly anticipate the strides the research community will make in this pursuit. Through Safurai-Csharp, we have showcased a step forward in this arena; however, the road ahead is long and full of fascinating possibilities and challenges. 

If you are interested in trying our model, here are the links to our Hugging Face page:
\\
Normal Version: \url{https://huggingface.co/Safurai/Safurai-Csharp-34B}
\\
GGUF Quantization: \url{https://huggingface.co/Safurai/Safurai-Csharp-34B-GGUF}
\\
AWQ Quantization: \url{https://huggingface.co/Safurai/Safurai-Csharp-34B-AWQ}

\newpage
\printbibliography

\end{document}